# A dancing bear, a colleague, or a sharpened toolbox? The cautious adoption of generative AI technologies in digital humanities research

Rongqian Ma[1*], Meredith Dedema[1], Andrew Cox[2]

[1] Luddy School of Informatics, Computing, and Engineering, Indiana University Bloomington, USA
[2] School of Information, Journalism, and Communication, University of Sheffield, UK

[*]Corresponding author

**Abstract**

The advent of generative artificial intelligence (GenAI) technologies is changing the research landscape and potentially has significant implications for Digital Humanities (DH)–a field inherently intertwined with technologies. This article investigates how DH scholars adopt and critically evaluate GenAI technologies for their research. Drawing on 76 responses collected from an international survey and 15 semi-structured interviews with DH scholars, we explored their rationales for adopting GenAI tools in research, identified the specific practices of GenAI tool use, and analyzed scholars' collective perceptions regarding the benefits, risks, and challenges of incorporating GenAI into their work. The results reveal that DH research communities hold divided opinions and differing imaginaries about the role of GenAI in DH scholarship. While scholars acknowledge the benefits of GenAI in enhancing research efficiency and enabling reskilling, many remain concerned about its potential to disrupt their intellectual identities. Situated within the history of DH and viewed through the lens of Actor-Network Theory, our findings suggest that the enrolment of GenAI technology, alongside other human and non-human actors within an integrated network, is gradually changing the field, though this transformation remains contested, shaped by ongoing negotiations. Our study is one of the first empirical analyses of this topic and has the potential to serve as a building block for future inquiries into the impact of GenAI on DH scholarship.

## 1. Introduction

Digital humanities (DH) is an established interdisciplinary field of inquiry attracting researchers from diverse areas both within the humanities and broadly from fields such as information studies and computer science (Luhmann & Burghardt, 2022; Weingart & Eichmann-Kalwara, 2017). Rooted in the integration of technological innovation and humanistic inquiry, DH, like other interdisciplinary fields such as computational social science or computational biology, has evolved closely with technological advances. Although its disciplinary boundaries remain fluid and amorphous, a shared methodological emphasis has unified diverse DH communities and shaped the field's core identity (Deegan, 2013).

Recent advances in artificial intelligence (AI) have accelerated this evolution. AI-based tools are reshaping DH by enabling large-scale analysis and revealing new interpretive possibilities across literature, history, and cultural heritage (Berry, 2022). Techniques such as natural language processing, topic modeling, stylometry, and geospatial analysis help scholars uncover patterns, attribute authorship, and visualize complex historical and textual data. These tools extend researchers' capacity to engage with vast datasets and derive insights that might remain hidden using traditional methods (Berry, 2022; Gefen et al., 2021). At the same time, DH scholars raise concerns about data politics, algorithmic transparency, and the interpretive limits of machine analysis in their work, including AI-driven research (Cottom, 2016; Schmidt, 2016; Underwood, 2025).

Over the past few years, the emergence of and rapid developments in generative AI (GenAI) models have created new opportunities to further rethink the relationships between humanistic inquiries and digital technologies. Foundational advances in deep learning during the 2010s, such as Generative Adversarial Networks (2014) and transformer models (2017), laid the groundwork for this shift. By the early 2020s, models like GPT-3 demonstrated strong language generation capabilities, while diffusion models enabled high-quality image synthesis. The release of ChatGPT in November 2022 marked a turning point in public and academic engagement with GenAI. As of March 2024, a growing ecosystem of GenAI tools, including GPT-4, DALL-E 3, Midjourney, Adobe Firefly, Gemini, and Stable Diffusion 3, reflects the field's rapid evolution. Compared to previous AI models, GenAI models learn from large datasets to generate original content, enabling broader applications in science, education, and communication (Marchese, 2022; Mitchell, 2023; Doron et al., 2025; Guo, 2024; Zhu & Molnar, 2025).

GenAI holds significant potential for impacting the arts and humanities, where deep interpretation, creative expression, and the distinctiveness of human insight are central values. Yet, contrasting perspectives persist on how GenAI should be treated in humanities research, particularly regarding its role in generating research content, its appropriate use, and its scholarly value (Gefen et al., 2021; Rane, 2023). Given this ongoing and contested discussions, it remains unclear whether GenAI is currently transforming, or will ultimately transform, DH as a field. This paper contributes to this conversation by presenting one of the first empirical studies examining how DH scholars engage with GenAI in their research. By exploring how these scholars critically and cautiously adopt GenAI, we aim to shed light on the complex and evolving dynamics of DH knowledge production in an age of emerging AI technologies.

Specifically, we investigate three major research questions:

1. What are the rationales behind the adoption of GenAI in DH communities?
2. How do DH scholars use GenAI technologies to support tasks in their research?
3. How do DH scholars perceive and critically evaluate the benefits, risks, as well as barriers of applying GenAI in DH research?

To answer the research questions in this early phase of GenAI adoption in DH, we used a combination of an online survey and semi-structured interviews to gather first-hand accounts and evidence from members of DH research communities. Analyzing the results from our survey and interviews through the theoretical lens of Actor-Network Theory (ANT), we highlight the dynamic and negotiated processes of enrolment, translation, and negotiation between human actors (e.g., individual DH researchers, their peers, and students) and nonhuman actors (e.g., GenAI, DH research tasks), which together may have the potential to transform DH as a field over the long term.

Grounded in the DH domain, this study advances research in information science by addressing three of its foundational concerns—the production of scholarly knowledge, scholarly communication, and the evolving nature of academic disciplines. DH has emerged as an area of interest within the broader field of information studies, owing to its interdisciplinary integration of humanities inquiry and digital technologies (Walsh et al., 2022; Zeng et al., 2022). In this study, we use DH as a domain to examine how GenAI technologies are negotiated and domesticated within scholarly information practices. By drawing on and extending theoretical insights from ANT, the study offers a domain-specific analysis that contributes to broader information science conversations about how emerging technologies are enrolled, negotiated, translated, and ultimately embedded in scholarly work and communities.

## 2. Literature Review

### 2.1 GenAI use in Research

The literature shows GenAI's potential in supporting various activities in scientific research, ranging from initial ideation, literature reviewing, data collection, and transcription, to data analysis and academic writing (Van Noorden & Perkel, 2023). Among scientific research communities, a significant number of researchers deem GenAI technologies either "useful" or "essential" (Van Noorden & Perkel, 2023). For example, Andersen et al. (2025) identified 32 specific GenAI use cases across five stages of the research process: idea generation, research design, data collection, data analysis, and writing/reporting. It is suggested that in the long term, involving GenAI in these stages of research may enhance productivity by shifting researchers' efforts away from repetitive, low-level tasks toward work that requires critical thinking and creativity, which may potentially transform the landscape of knowledge work (Buruk, 2023). Another empirical study highlighted the possibility of GenAI to enhance qualitative research by supporting coding tasks and contributing to code generation and refinement during the initial open coding phase of the grounded theory analysis (Sinha et al., 2024).

Filimonovic et al. (2025) used a large author-level panel dataset from the Scopus database and applied a difference-in-differences (DiD) approach to investigate how the adoption of GenAI, such as ChatGPT, impacts academic productivity and research quality in the social sciences and psychology. Their analysis finds that GenAI users significantly increase their publication output—by 15% in 2023 and 36% in 2024—and experience modest gains in publication quality, as reflected in higher journal impact factors. These effects are strongest among early-career researchers, those in technically demanding fields, and scholars from non-native English-speaking countries, suggesting GenAI helps lower barriers related to language and technical skill. Andersen et al. (2025) investigated how researchers at Danish universities use GenAI in their work and how they perceive its implications for research integrity. The study employed a nationwide survey in early 2024, collecting 2,534 complete responses. The main findings reveal that GenAI is widely but selectively used—most positively for language editing and data analysis, and more cautiously or critically for tasks like experimental design or peer review. Junior researchers and those in technical fields were more likely to adopt GenAI tools and assess their use positively, whereas humanities scholars and qualitative researchers tended to be more skeptical. Importantly, while actual use remains modest, favorable assessments outpace reported usage, suggesting that limited awareness, skills, or institutional support, rather than ethical concerns, may be barriers to broader adoption.

However, integrating GenAI in research is not without concerns (Delios et al., 2025). As Marchese (2022) observes, AI technologies do not have a "true understanding" of knowledge in a human sense, which contributes to issues such as hallucinations and misinformation propagation. The level of accuracy of GenAI is a critical problem for research. Integrating GenAI in research methodologies also risks producing non-replicable results, diminishing the transparency and quality of research (Dwivedi et al., 2023). Beyond discussions from such an outcome-based, teleological perspective, concerns about GenAI's value in research also extend to a deontological, process-oriented perspective (Schlagwein & Willcocks, 2023). Intellectual property and plagiarism represent two ethical dilemmas exacerbated by GenAI involvement: With AI-generated content coming into question, researchers worry that it might compromise the originality and authenticity of academic contributions (Lund et al., 2023; Mitchell, 2023).

Over-reliance on GenAI for recognizing patterns, for example, may detract researchers from gaining a profound understanding of the subjects at hand (Khlaif et al., 2023). The absence of nuanced judgment and sensitivity to context in AI-generated materials could also lead to a degradation of critical analysis and sophisticated understanding, which are essential elements of scholarly work. These considerations raise the question of whether GenAI's lack of human qualities compromises the processual integrity valued by deontological (process-oriented) ethics, or whether the emphasis should instead shift to beneficial outcomes—such as greater efficiency and knowledge enhancement—aligned with teleological ethics (Schlagwein & Willcocks, 2023).

Beyond the impact of GenAI in the scientific landscape, there are societal impacts that researchers also relate to, for example, GenAI could lead to cultural centralization, marginalizing diverse languages, traditions, and knowledge systems (UNESCO, 2021). Moreover, GenAI may deepen divides between scholars in technologically advanced countries and low- and middle-income countries in terms of access to AI benefits and representation in AI development (Wang et al., 2024). Finally, scholars are concerned with the environmental costs of GenAI technologies, such as high energy consumption, greenhouse gas emissions, and rare metal extraction (Berthelot et al., 2025).

### 2.2 Application of AI in DH

Researchers in the field of digital humanities (DH) have long leveraged computational methods to advance the study of human culture, history, and literature, beginning with the field's inception in the 1950s. In recent decades, the integration of AI technologies has significantly expanded these efforts. By facilitating the analysis of large datasets and textual corpora with AI, scholars have been able to uncover patterns and insights not readily visible through traditional methods (Berry, 2022; Gefen et al., 2021). For instance, natural language processing (NLP) is employed to analyze vast amounts of text, revealing trends, sentiments, and thematic connections across historical documents and literary works at a scale not possible for the individual researcher (Jockers, 2013; Moretti, 2013). This shift towards computational analysis and "distant reading" of texts has complemented traditional interpretive methods, such as close reading. However, this enrichment relies on scholars' critical and reflective engagement with these tools, highlighting that while AI integration has created new opportunities in DH, it also brings significant challenges (Underwood, 2019, 2025).

On the bright side, advances in AI-based technology are reshaping how labor is conceptualized and practiced in DH. The blurring of disciplinary boundaries in DH, as Underwood (2016) argued, enables interdisciplinary collaboration and allows for the use of computational, distant reading methods to address more complex research questions. Further, Smithies (2017) argued that AI has become a central technology of the "digital modern," where the heavy reliance on automation will potentially offer new opportunities for innovation and collaboration in DH.

On the darker side, concerns have emerged about the use of AI-based tools in DH, particularly regarding the politics of data and the transformative effects of algorithmic systems. For example, Cottom (2016) notes that as we move to large-scale textual and data-driven analysis, the very affordances of "big data" can clash with the nuanced, context-sensitive questions at the heart of humanities and social science inquiry because, "with scale come questions that are important for how we identify, understand, and analyze textual data," but those questions often reveal that the categories and structures imposed by datasets may not align with critical theoretical concerns—forcing scholars to negotiate and adapt their frameworks to the constraints of the data. Schmidt (2016) further pointed out the danger of misusing algorithms: when algorithms are treated as black boxes with unexamined assumptions, they can produce misleading results that obscure or erase meaningful distinctions, ultimately undermining the interpretive goals of the humanities.

Building upon this long history of AI adoption in DH, recent rise of GenAI technologies may be yet another turning point for DH research. Existing literature has demonstrated the use of GenAI technologies to assist with a variety of DH tasks, such as data annotation, translation and transcription, metadata enhancement, collection curation, and teaching of the humanities (Colutto et al., 2019; Guo, 2024; Nockels et al., 2022). For example, Karjus (2023) discussed the potential of integrating LLMs in humanities research, proposing a machine-assisted mixed methods framework and suggesting its application in diverse linguistic and disciplinary contexts. Rane (2023) further explored the potential impact of ChatGPT in the arts and humanities, outlining how this AI tool could serve as a creative collaborator that aids artists and scholars in ideation, data analysis, and literature synthesis. The commitment of humanities research to deep, critical engagement with complex cultural narratives—a task traditionally reliant on the intricate subtleties of human cognition and interpretive capacity—can potentially be further reshaped with GenAI's entry into this sphere, posing questions about the balance between algorithmic engagement and the rich, often intangible insights derived from human analysis.

However, scholars have also begun to speculate about the impact of integrating GenAI into the arts and humanities, raising questions related to data privacy, intellectual property, originality and authorship, and broader scholarly paradigms. Rane (2023), for instance, points to data privacy and algorithmic bias as possible ethical concerns, suggesting that GenAI systems like ChatGPT could perpetuate and even amplify existing societal biases embedded in their training data. Authors also theorize about how GenAI might complicate traditional notions of authorship and originality, especially in contexts where the creative process lacks the intentionality and emotional depth associated with human scholarship (Oxtoby et al., 2023; Rane, 2023). In this vein, scholars have proposed that the growing presence of GenAI may prompt a rethinking of long-standing models of scholarly production—from the figure of the solitary researcher to collaborative or co-intelligent partnerships. These forward-looking reflections underscore the importance of anticipating potential ethical and institutional challenges as GenAI continues to evolve.

Despite these developments, there is a noted scarcity of systematic empirical research examining how emerging GenAI tools are currently perceived, used, and evaluated in DH research, indicating a crucial gap in the field that warrants further investigation. Our present study aims to bridge this gap by surveying DH scholars on their use of these emerging technologies, as well as their collective perceptions regarding the adoption of GenAI in research processes.

### 2.3 The Actor-Network Theory and the Role of Technology in DH

While many theories in science and technology studies seek to explain how scientific advances occur and disciplines evolve (Merton, 1973; Kuhn, 1962), Bruno Latour's (1988, 1992, 1996, 2005) Actor-Network Theory (ANT) offers a particularly effective framework for understanding the co-constitutive roles of technology, humans, knowledge, and practice in these dynamic processes. Rather than viewing science as a purely human or linear endeavor, ANT reconceptualized it as the product of shifting networks composed of both human and non-human actors (Latour, 1992, 2005). Technologies, instruments, texts, organisms, and people all operate as actors whose agency is distributed and relational (Murdoch, 1997). And therefore, scientific disciplines emerge and evolve not through discrete paradigm shifts alone, but through the continual reassembly of heterogeneous networks—networks that are enacted and maintained through processes of enrolment and translation, in which actors negotiate, align, and redefine one another's roles and interests.

In ANT, *enrolment* refers to the dynamic process by which actors are recruited and integrated into a network, taking on specific roles and responsibilities. *Translation*, as defined by Latour (1988, 1996), describes how actors with differing degrees of power and interest negotiate alignment in order to form and stabilize a network. Latour outlines five modes of translation: weaker actors may align their goals with stronger ones; stronger actors may adjust to support weaker ones; actors may compromise or detour temporarily from their aims; they may reinterpret goals altogether; or they may become *obligatory passage points*—indispensable intermediaries that others must pass through to achieve their own ends (Luo, 2020, Chapter 1, pp. 8–35). These modes reveal the strategic, relational nature of network formation, in which actors continuously reshape relationships and reconfigure interests to enable collective action.

ANT emphasizes the agency of technology and other nonhuman elements as integral actors that actively contribute to the formation and transformation of networks. For example, printing presses and visualization instruments function as inscription devices that stabilize knowledge and redefine intellectual labor, governance, and power (Latour, 1990). As shown in a few empirical studies, technologies such as speech recognition systems in clinical settings or institutional repositories in academic publishing are not adopted in isolation but become entangled in professional workflows, institutional routines, and disciplinary norms (Morland & Pettersen, 2018; Kennan & Cecez-Kecmanovic, 2007). Through such entanglements, technologies are not only adapted to fit existing practices but also reconfigure those very practices, illustrating the recursive dynamics of sociotechnical co-evolution.

We argue that the ANT framework offers a productive perspective to carry out the study. First, ANT's particular emphasis on nonhuman agency makes it especially appropriate and meaningful for thinking about GenAI, a form of technology increasingly perceived as exhibiting human-like capabilities. As the AI literature increasingly discusses AI's agentic behaviors, such as the ability to use tools, make decisions, and complete tasks (Murugesan, 2025), the ANT notion of non-human agency inspires us to reflect on whether GenAI—unlike previous generations of digital technologies—may be transforming, or at least has the potential to transform**,** DH in significant and fundamental ways. Moreover, ANT's focus on the enrolment and translation of sociotechnical networks is well-suited to the current moment, as GenAI remains in an early stage of adoption within DH—its roles, affordances, and implications are all still actively unfolding. By identifying the diverse human and nonhuman actors involved, we can better understand how GenAI and DH research interact, negotiate, and contribute to each other.

As a field shaped by layered technological, institutional, and intellectual entanglements, the development of DH exemplifies the socio-material assemblages that ANT seeks to theorize. Its formation, as Dalbello (2011) argues, reflects a cumulative history of material, technological, and methodological transformations. The emergence of DH is not a singular event or paradigm shift, but rather, the outcome of an ongoing negotiation among researchers, technologies, methods, infrastructures, and disciplinary values. Empirical studies show that many humanities scholars approach technology use with a cautious pragmatism, adopting tools only when they perceive a clear benefit to their research workflows (Given & Willson, 2018). Technologies that support large-scale text analysis, for instance, align with longstanding scholarly practices in the humanities, even as they reshape those practices in subtle but consequential ways. These acts of selection, customization, and integration of tools are acts of translation–or in other words, negotiations, in which actors' identities, purposes, and affiliations are redefined in the actor-network. Moreover, many researchers are not simply users of digital tools but also toolmakers and infrastructural thinkers. Tool development itself is often framed as a form of scholarly inquiry, challenging traditional boundaries between research and technical labor. This dual identity–researcher and developer–complicates any simple account of technological adoption as a one-way process. The conceptual ecology of DH, as elaborated by Poole (2017), further underscores the multidimensional roles that technology plays in the field. Drawing on Svensson (2009, 2010), Poole outlines how information technology in DH can function as a tool, an object of study, a medium, a laboratory, or even a platform for activism. This plurality resonates with ANT's refusal to assign fixed roles or essences to actors; instead, the significance of a given tool or practice emerges through its embeddedness in situated networks of use, meaning, and power.

In this sense, the ANT principle that knowledge practices are enacted through networked relations between human and non-human actors may have considerable explanatory power when considering the impact of GenAI on DH. In the following sections, we draw on this lens of ANT to examine the gradual, negotiated process of GenAI adoption within development of DH, particularly discussing whether GenAI is transforming DH, and if it is, in which ways, and how.

## 3. Data and Methods

This study employed a sequential mixed-methods design, combining a survey screener with semi-structured interviews as its primary research methods. The survey screener allows for broad engagement with the DH community and supports effective participant recruitment, while the semi-structured interviews enable the collection of rich, in-depth insights into participants' thoughts, experiences, and perceptions. These interviews also help elicit concrete case studies and examples of GenAI use in research (Creswell & Creswell, 2023). Integrating breadth and depth, this approach captures a wide range of perspectives on the adoption of GenAI in DH, an area characterized by rapid development, ongoing debate, and potential divergence within the field. This study has been approved by the Indiana University Institutional Review Board (IRB), under study #21946.

***Survey Design and Distribution.*** The survey screener was built in Qualtrics and distributed via DH and information-science listservs (e.g., DHSI, ADHO) as well as the authors' personal social channels (e.g., X) to recruit participants. The survey included questions about DH scholars' current practices and their perceived benefits, challenges, and barriers related to the use of generative AI tools in research (see supplementary material for the full list of survey questions). Several items were adapted from Van Noorden and Perkel's (2023) study according to the DH context, which surveyed over 40,000 scientists on their views regarding the integration of GenAI into scientific research. We distributed the survey between February 18 and March 20, 2024, and received 151 responses, including 76 fully completed ones in which participants answered all questions up to the designated exit point based on their eligibility (see Figure 1). Of the 76 completed responses, 59 participants self-identified as DH scholars, and among them, 37 reported having used GenAI in their research. The remaining 22 DH scholars, who had not adopted GenAI, shared their reasons for non-adoption before exiting the survey.

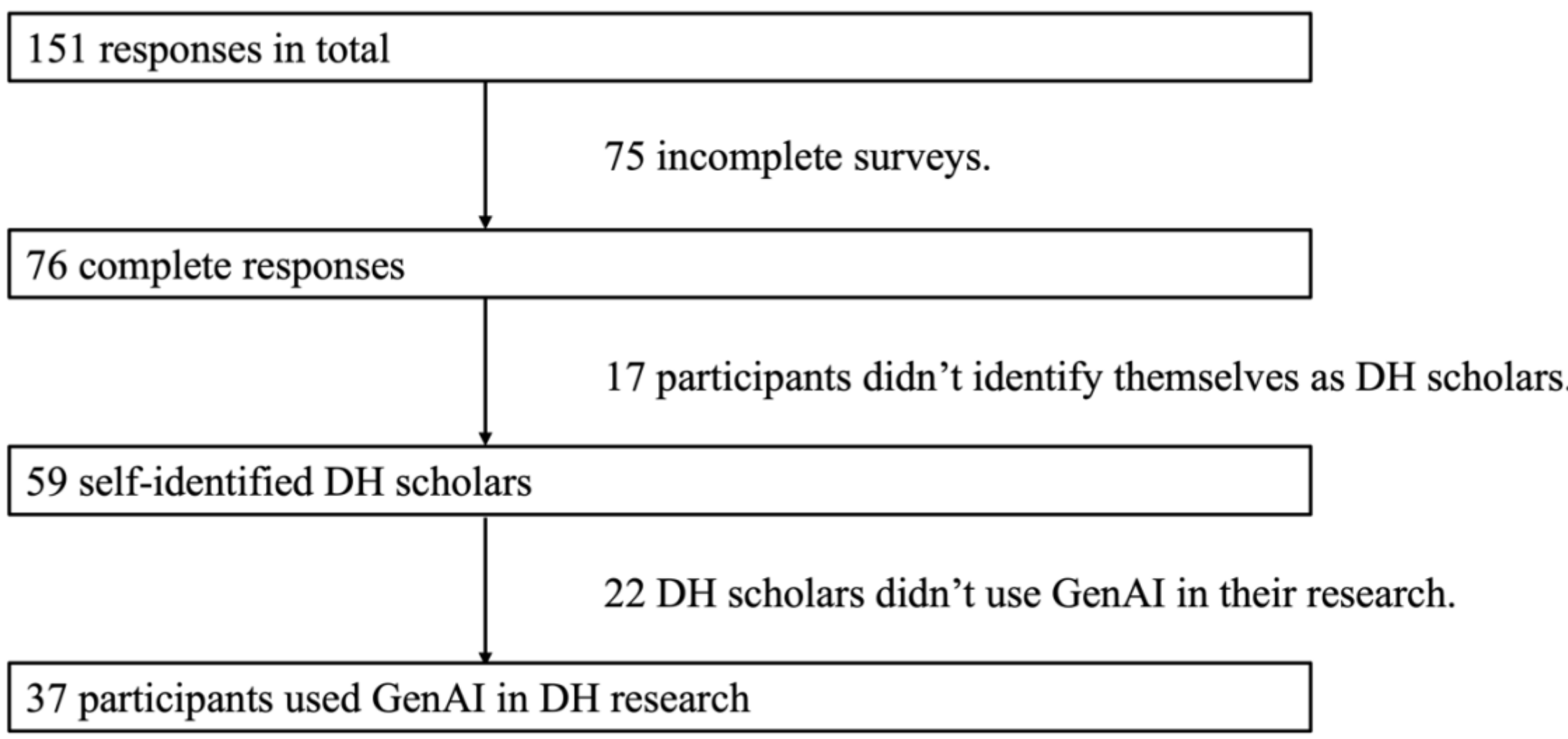

Figure 1. Results of the survey screener.

Participants reported working in a diverse range of DH domains and subfields, such as digital history (n=5), spatial analysis (n=3), knowledge organization (n=3), languages, literature, and linguistics (n=3), digital archives (n=2), storytelling (n=2), and library and information science (n=2).[1] These DH scholars identified as professors, research fellows, lecturers, and graduate students and provided broad international perspectives, coming from the U.S. (14), Europe (8), Canada (5), Middle East (2), Mexico (2), Africa (1), China (1), and the U.K. (1), with three additional unknown locations. Participants' experience in digital humanities ranged from one to 28 years, with an average of eight years.

[1] The examples listed here include only categories with at least two participants. One of the authors grouped and labeled these fields based on the participants' self-described research areas.

***Semi-structured Interview.*** From the 37 responses (P1–P37) from DH scholars, we identified those who expressed interest in participating in follow-up semi-structured interviews for deeper conversations. We sent out interview invitations to these respondents and secured 15 interviews. The interviews were conducted between February and August 2024. Each interview lasted for about 45-60 minutes and was audio-recorded for transcription purposes. As shown in Table 1, the interview participant pool includes 15 DH scholars representing a range of academic ranks, disciplinary backgrounds, geographic regions, and levels of experience in DH research.

Table 1. Information about the interview participants (identified by their original IDs from the pool of 37 invited individuals).

| ID | Title | Domain[2] | DH Career Age | Region |
|---|---|---|---|---|
| P02 | Researcher | Digital infrastructure, digital humanities | 18 | Europe |
| P03 | Associate Professor | Historical GIS, web archives, public history, immersive technology, linked open data | 15+ | Canada |
| P06 | Researcher | Digital History | 5 | Europe |
| P07 | Professor | English; User Studies | 28 | UK |
| P08 | Director | History | 30 | US |
| P11 | Graduate Student | Digital history, multilingual digital humanities | 7 | US |
| P12 | Lecturer | Rhetoric and composition, food and drink rhetoric | 2 | US |
| P13 | Associate Professor | English; Text mining and image analytics in digitized newspaper collections | 17 | US |
| P14 | Assistant Professor | Organized disinformation operations and media literacy | 10 | Mexico |
| P19 | Research Associate | Spanish literature | 10 | Europe |
| P21 | Assistant Professor | Cultural heritage | 3 | China |
| P22 | Research Fellow | Digital history | 10+ | US |
| P23 | Project Manager | Digital Humanities | 5+ | US |
| P24 | PhD Student | Cultural heritage | 2 | China |
| P27 | Assistant Professor | Information science | 5+ | US |

[2] All the domains are described using scholars' self-defined terms.

Our interview protocol (see supplementary materials) was structured as three sections. The first section gathered participants' demographic information, focusing on their institutional affiliations and academic positions. In the second section, we asked participants to reflect on their own research as case studies and to describe how they use and discuss GenAI at various stages of the research process. We defined these research stages as follows:

- **Research conception or preparation:** e.g., conducting literature searches and reviews, brainstorming research questions;
- **Data collection and construction:** e.g., searching for data sources, primary sources or archives; checking on the updates of targeted datasets; working on codes for data collection; conducting fieldwork;
- **Data analysis:** e.g., basic descriptive analysis, collaborative coding, content analysis, running and testing computational models, data visualization;
- **Research outcome communication:** e.g., manuscript writing, editing, publishing, or presenting; delivering digital projects.

Based on participants' descriptions of their use cases, we asked them to reflect on what motivated their adoption of GenAI, how they incorporated it into their research, and how they evaluated its advantages, benefits, as well as the challenges and concerns associated with its use.

***Interview coding and analysis.*** Given the emerging nature of this topic and the lack of an established analytical framework in the existing literature, we adopted a grounded theory approach combined with deductive coding to develop interpretations and identify major themes from the data (Charmaz, 2006). The first two authors began with an open coding round of the interview transcripts using NVivo and collaboratively developed an initial codebook. After discussing and refining the codebook, the authors proceeded with a second round of axial coding. In this stage, the authors divided the transcripts and independently annotated their assigned portions using the revised codebook (see supplementary materials).

## 4. Findings

### 4.1 Rationales behind GenAI Adoption in DH Research

Our survey and interviews reveal that the main motivations behind researchers' adoption of the tool are based on GenAI's perceived benefits, as well as the effect of community influence. Specifically, two major benefits are seen by DH scholars: 1) improving research and work efficiency, and 2) potentially enabling more fundamental transformations in research. On one hand, as the survey data shows (Figure 2), many researchers adopt GenAI to streamline routine tasks and enhance productivity. Sixty-two percent of participants reported that GenAI tools "make coding easier and faster," while 41% cited their utility in "speeding administrative duties" and "summarizing other research to save time reading it" (41%). Additionally, the survey also implies that these tools are seen as supporting greater equity in scholarly work, with 62% highlighting their role in "helping researchers without English as a first language (through editing or translation)" and 14% acknowledging that they can "help write manuscripts faster."

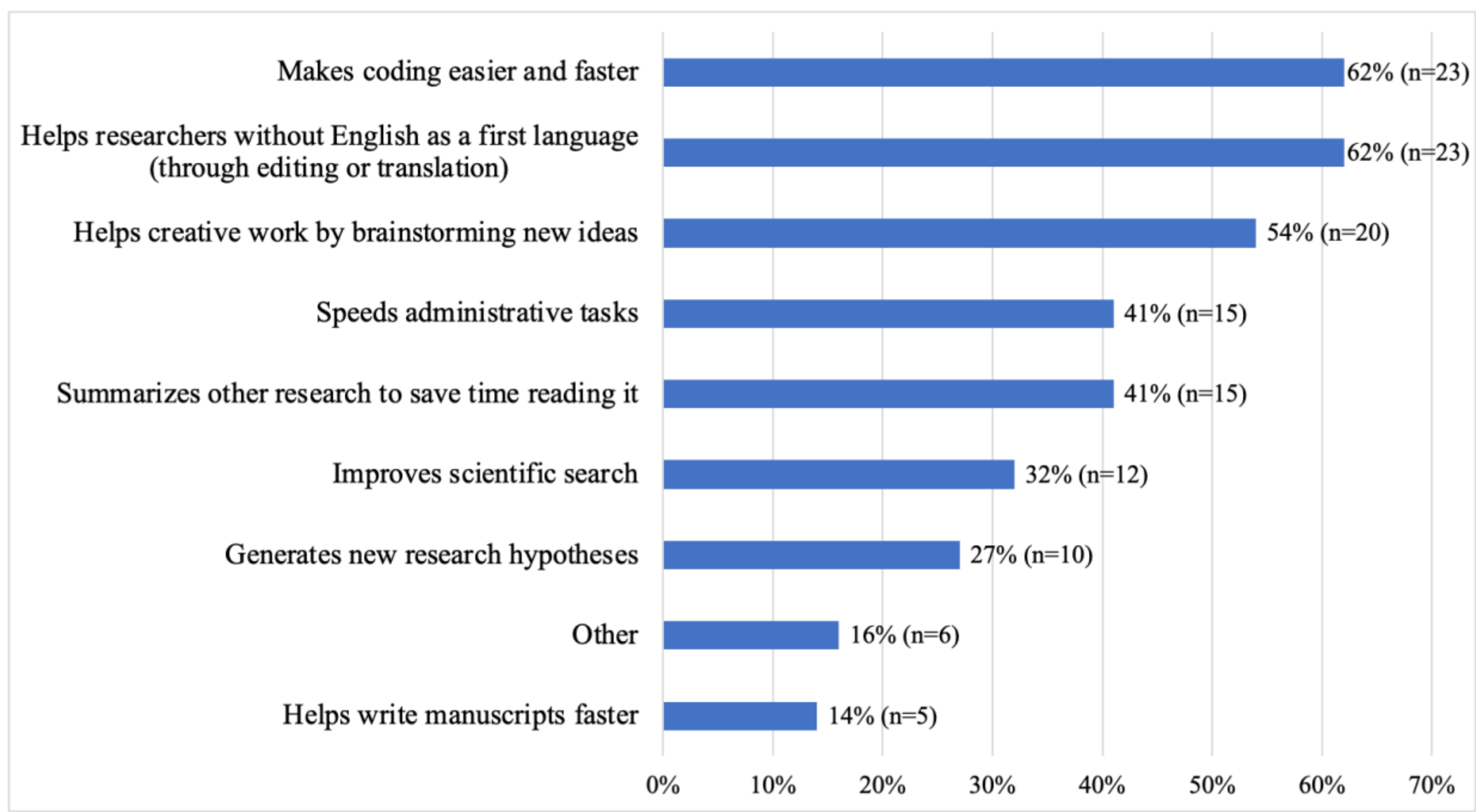

Figure 2. Benefit of using GenAI in research based on survey responses (total number N=37).

Consistent with the survey results, interview data also show perceived benefits of GenAI in increasing efficiency. Across various stages of research, from data processing to coding to search and synthesis, GenAI is seen as enabling researchers to save time, reduce labor-intensive tasks, and redirect their focus toward higher-level intellectual work. For example, GenAI is applied to drastically shorten the time required to go through large volumes of archival materials and primary sources. As P2 noted, working with a five-million-page corpus of Dutch East India Company records,

> "If you search for something… and you get 30 hits… 20 pages each… that's 600 pages to read in the Dutch original. But if you could ask GPT to summarize and suggest which ones are the most relevant… then maybe you only have 3 documents to read."

This allows researchers to prioritize what's worth close reading, making large corpora more navigable and manageable. One researcher also noted how ChatGPT transformed their programming workflow:

> "I could go from being a lower intermediate [programmer] to maybe like an upper intermediate coder really quickly… I can think at a higher level of what I want the algorithm to do… and not have to worry about where the commas go." (P3)

On the other hand, DH scholars also pointed to GenAI's potential to transform research more fundamentally. For example, survey participants noted its value in creative and intellectual work (Figure 2), including "helping creative work by brainstorming new ideas" (54%) and "generating new research hypotheses" (27%). Another major benefit of incorporating GenAI into research, as reported by DH scholars, is its role in reskilling. GenAI has been widely embraced as a tool for accelerating the acquisition of computational skills and supporting project implementation. As P8 explained, using GenAI as an entry point into unfamiliar technical domains enabled him to take on tasks he might have otherwise avoided due to a lack of formal expertise, such as data analysis and visualization:

> "I do expect to use it for things like data analysis…. I don't know how to do coding and I don't know data visualization that well, so I think that it might be able to help with that sort of thing." (P8)

Beyond perceived benefits, a third rationale for GenAI adoption among DH scholars, as suggested by the interview data, is shaped less by perceived utility and more by external influences and community context. For instance, P7 began experimenting with GenAI after a physics colleague introduced it during a research collaboration, while P8 and P12 credited institutional workshops for sparking their initial interest. These accounts highlight the importance of community-level initiatives, peer influence, and institutional support in fostering meaningful engagement with GenAI, showing that adoption is as much a social process as it is a technical or intellectual one.

### 4.2 DH scholars' use of GenAI in research: Practices

The perceived benefits of GenAI, along with external community influences, have likely contributed to its adoption in DH research practices. Our survey and interview data show that GenAI is used throughout the DH research lifecycle to engage in a variety of tasks. As seen in Figure 3, the most common application is brainstorming research questions and ideas—a creative and exploratory activity reported by 62% of respondents. This aligns with our interview data, which highlights the generative potential of these tools in the earliest phases of inquiry.

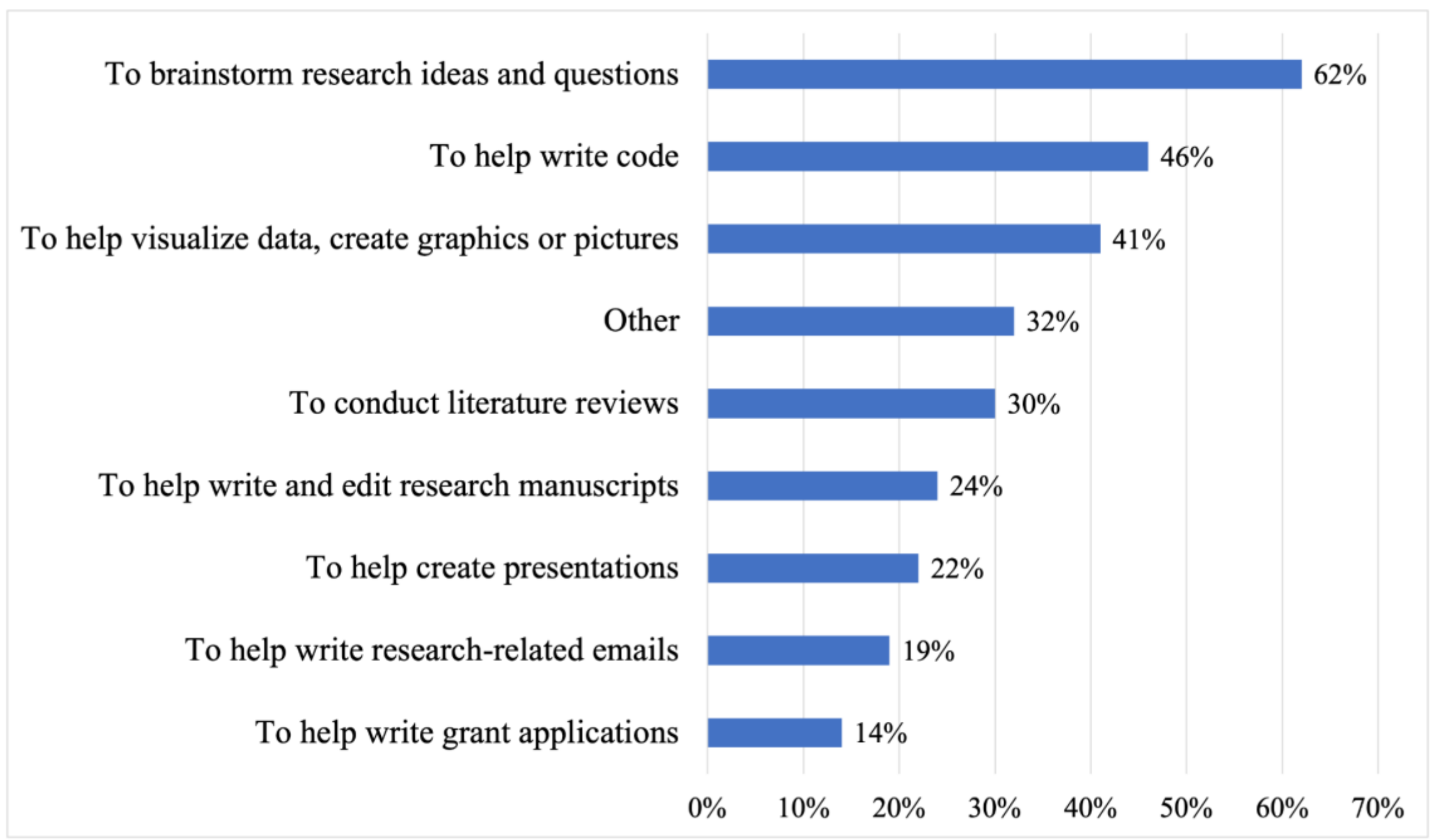

Figure 3. DH scholars' purpose of using GenAI in research based on survey responses (total number N=37).

In data collection, scholars are applying GenAI to longstanding challenges such as Optical Character Recognition (OCR) correction, metadata generation, and even synthetic dataset creation. As P3 described:

> "[For] nineteenth-century sources OCR is a hot mess... But you can get generative AI… you can cut and paste the bad OCR in the ChatGPT and give it clear instructions... And it will give you a much cleaner, much closer to the original return."

P22 similarly leveraged multimodal GenAI to transcribe handwritten manuscripts:

> "We originally used a general OCR model... [but switched to] generative AI... especially the multimodal AI, rather than just traditional machine learning."

Others have creatively generated synthetic data to fill resource gaps. As P27 noted:

> "We generated song lyrics using ChatGPT... lyrics are copyright protected... So we just built a lyrics dataset upon the Million Song Dataset... ChatGPT generated lyrics using the list of words... it makes a certain mood, genre, and style."

In data analysis, interviewees mentioned uses such as translation, classification, qualitative coding, entity recognition, and retrieval-augmented generation (RAG). P6 explained how GenAI supports text normalization and named entity recognition:

> "I used ChatGPT to correct and normalize the transcriptions provided by Transkribus, analyze the texts and recognize all entities (persons mentioned, dates, places, events narrated)."

P11 described evaluating multiple models to improve translations of Ottoman Turkish texts: "We tested different GenAI models (GPT-4, Gemini) and fine-tuned a model to get better outcomes between Ottoman Turkish and English."

Coding and computational tasks also emerged as a common theme. GenAI served as a coding assistant, Q&A partner, or coach, supporting both experienced programmers and those new to coding. P3 shared: "It allows [me] to do things (i.e., coding and programming) that [I don't] do every day." P21, however, emphasized that some coding background is still necessary to fully benefit: "You do need certain knowledge and background in coding... to maximize the benefit of GenAI as a 'coach.'"

In writing, GenAI tools are mostly used for light editing, proofreading, and improving fluency. Non-native English speakers particularly noted the benefit of having a tool that can refine their work without relying on human editors (e.g., P21). And P11 added: "It gives feedback on sentences, so I don't need to get the writings edited [by a professional editor]." While less frequently used for formal communication, some survey respondents do apply GenAI for tasks like composing grant proposals or research-related emails, albeit cautiously.

Overall, both survey and interview data reveal a consistent pattern: GenAI tools are primarily favored for internal, exploratory tasks that enhance creativity and individual productivity. Their use in public-facing scholarly work remains limited–likely due to ongoing concerns about accuracy, originality, and institutional disclosure requirements. Existing research has pointed to the phenomenon of an "AI disclosure penalty," suggesting that revealing the use of AI in scholarly outputs may lead to negative judgments (Reif et al., 2025). At the same time, attitudes within scholarly communities toward the use of GenAI in academic work remain varied and unsettled (Kwon, 2025).

## 4.3 Perceptions of GenAI among DH Communities

### *4.3.1 Risks and concerns among scholars of using GenAI in DH research*

However, the adoption of GenAI in research is not without concerns. Survey data (Figure 4) show that 78% of DH scholars (n = 29) anticipate "mistakes or inaccuracies" in AI-generated texts, and 68% (n = 25) express concern about the spread of misinformation. In interviews, participants described GenAI systems as opaque and prone to hallucinations. As one interviewee noted, "When it gives an answer, you don't know if it's drawing on its background knowledge… or from the 14 speeches you told it to draw from" (P2). Another remarked, "It smooths over things it doesn't know… It just kind of confidently presents a result that is a lot less certain than it seems" (P8). Such behavior undermines the evidentiary rigor and source-critical standards that are foundational to humanities scholarship.

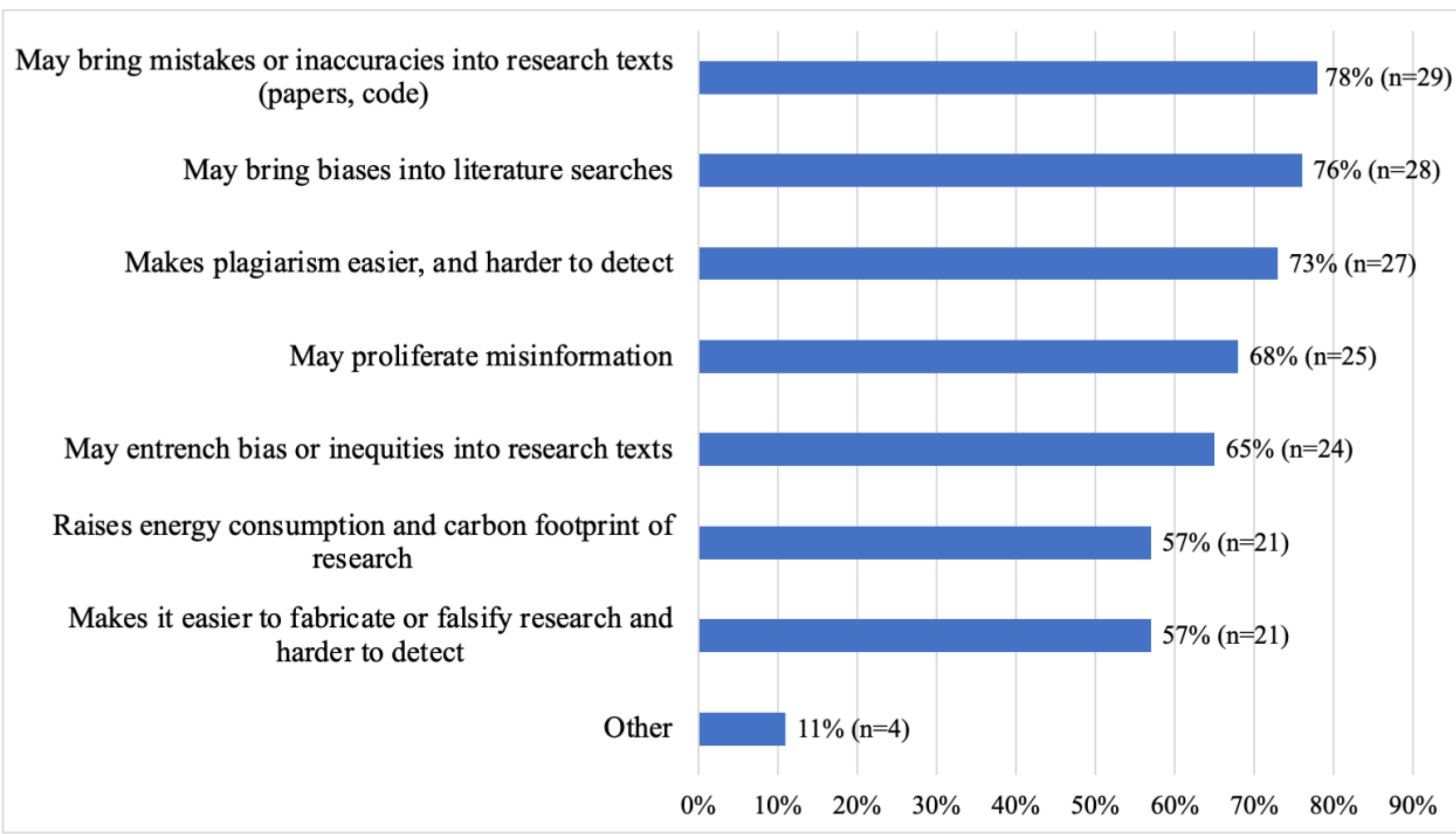

Figure 4. The risks of using GenAI in research among DH scholars based on survey responses (total number N=37).

Concerns about accuracy are closely linked to worries about bias and inequity. Seventy-six percent of respondents believe GenAI could skew literature searches, and 65% anticipate biased textual outputs. Interviewees also pointed out that most training data are drawn from English-language, North American sources: "The datasets are predominantly North American, predominantly English… it creates a certain model that produces certain worldviews," explained P11. For those working with underrepresented languages or cultural contexts, these limitations threaten to misrepresent, or even erase, critical nuances.

Research integrity and authorship present additional challenges. Seventy-three percent of scholars indicated that GenAI “makes plagiarism easier, and harder to detect,” and 57% fear it facilitates fabrication or falsification of results. One interviewee, P8, echoed this concern: “[GenAI] plagiarizes pretty badly… it will take large chunks of [sources] and just stick them in.” The same participant also described how AI contributions can blur the line between tool and co-author: “If you’re using something else to write for you, then that’s a coauthor as far as I’m concerned. But can it [ChatGPT] truly be a co-author?”

Beyond these epistemic and ethical issues, many DH researchers feel uncomfortable about GenAI’s impact on their professional identities While these concerns were not quantified in our survey, interviewees emphasized that reliance on AI risks bypassing the intellectual processes that make research meaningful: “Going through these processes makes you a better researcher… If everybody just uses AI… it’ll just start summarizing what everybody else has done,” warned P8. Others questioned the future role of academics when powerful AI tools exist: “Who can build models like ChatGPT? Professors in prestigious institutions cannot. Then what is our purpose?” (P27). In response, some participants argued for guidelines or regulations to protect creative labor: “For any creative work it has to be either prohibited, banned, or regulated… we are risking making those creative people poorer,” suggested P19.

Finally, environmental impact emerged as a significant concern among respondents. More than half (n = 21, 57%) highlighted the high energy consumption and carbon emissions associated with training and deploying large GenAI models. This concern aligns with broader societal anxieties about the environmental costs of AI and reflects increasing attention to sustainability in policy discussions, including those led by UNESCO (2021). Overall, the survey and interview findings reveal that DH scholars’ reservations about GenAI are multifaceted, encompassing technical reliability, algorithmic bias, research integrity, the value of human expertise, and environmental responsibility.

#### *4.3.2 Barriers to integrating GenAI in DH research*

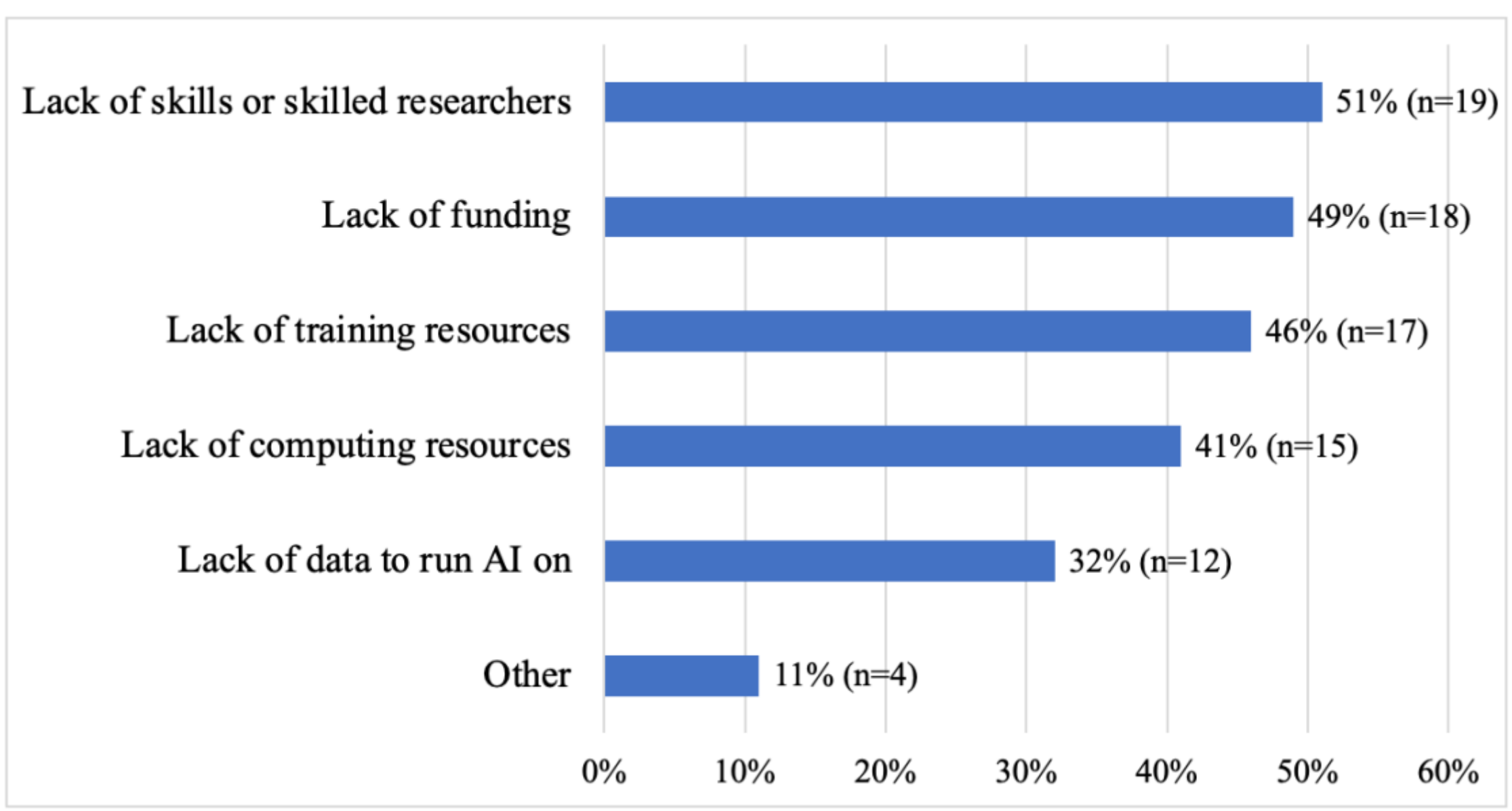

Figure 5. The barriers to using GenAI in DH research based on survey responses (total number N=37).

Besides the perceived risks, certain infrastructural barriers could prevent researchers from further integrating GenAI in research. Figure 5 illustrates that 51% (n=19) of scholars identified the "lack of skills or skilled researchers" as a principal barrier. One survey participant noted that their main reason for not using GenAI in research is "not hav(ing) anyone at (their) project or institution that is familiar enough with AI to assist us."

Related to the lack of AI skills, P12, discussed the difficulty of prompting skill development:

> "I know that it has some of these capabilities, but I myself am not very experienced in that kind of research. So I don't know how to prompt it to do what I want it to do in that way." (P12)

This shows how a lack of familiarity with prompting techniques creates a barrier, especially for tasks involving data construction and analysis. In addition, a common misunderstanding of GenAI as a static Q&A system, according to P12, was another barrier for her (and her students) to effectively use GenAI in research:

> "[Students are] not as familiar with dialogue and conversing to think through ideas... They'll ask a question and if they incorrectly prompted it... they're like, 'well it doesn't work, it's broken.'" (P12)

Following the lack of skills, survey respondents also identified the "lack of funding" (n=18, 49%), "lack of training resources" (n=17, 46%), and "lack of computing resources" (n=15, 41%) as major barriers. One interviewee (P22) also mentioned that "[ChatGPT] API [is] very expensive." Finally, 32% (n=12) of the participants identified a "lack of data to run AI on" as a significant factor leading to the underutilization of GenAI. Within the "other" category (n=4, 11%), survey participants noted obstacles stemming from the GenAI tools' inherent limitations. For instance, one participant described ChatGPT as a "shapeless tool," pointing out the difficulty in determining its optimal use-case scenarios due to its broad capabilities. Another respondent emphasized the challenge presented by GenAI's "lack of integration with open-source software" which further impedes researchers from incorporating these tools in their work.

Beyond the above barriers, one interviewee also discussed interesting motivational and psychological barriers:

> "…when using GenAI, it's a lot of experimenting… and a lot of things won't work. Somebody has to be open to doing stuff that way. …I teach mostly freshmen… they struggle with using generative AI because they're afraid to experiment… They expect it to work one way, and when it doesn't, they think, 'well, it's useless,' instead of trying different things." (P12)

Users, especially students and novice users, often lack confidence and are reluctant to engage in the trial-and-error process that GenAI requires. Their fear of being wrong or making mistakes discourages experimentation, leading to abandonment of the tool when it does not yield immediate results. This mindset limits their ability to explore GenAI's potential in research.

#### *4.3.3 Responses to the challenges*

The potential risks associated with GenAI adoption have elicited a range of responses from DH scholars. While some have opted against its use due to concerns such as hallucination and misinformation, others have cautiously integrated GenAI into their work, developing coping strategies and solutions tailored to their specific research and pedagogical contexts.

For example, to address the primary risk of GenAI producing misinformation or hallucinated content, several participants emphasized the need for rigorous validation. Six interviewees emphasized the necessity of thoroughly verifying AI-generated outputs. As P8 noted: "I'm going back and forth with it to see if I might trust it. But that trust only comes after I verify everything it produces, repeatedly, until I feel confident. You really have to check it, annotate it, and make sure it's done correctly." P13 offered a similar approach in relation to programming tasks: "I asked for the code running the final version and executed it on my own data and system, just to ensure nothing was being altered in the process."

In addition to manual validation, some scholars are implementing formal evaluation frameworks. P23 described a comparative approach to assess large language models, particularly for complex multilingual tasks: "We use precision and recall—standard statistical measures—to evaluate performance. We test each model on texts with intricate logic and narrative structures before applying them in actual research." P27, drawing on her computer science background, noted that her technical expertise enables her to critically assess the accuracy and efficacy of AI-generated code.

Concerns over research integrity have also led scholars to reconsider and redesign existing practices. For instance, P8 restructured course assignments to integrate GenAI in a pedagogically transparent manner: "In one assignment, students used GenAI to produce a first draft, which they then critiqued. In another, they wrote an initial draft themselves and used GenAI to refine it." Similarly, P19 proposed the use of explicit AI-use disclaimers as a means of maintaining transparency and academic rigor in published work.

These findings suggest that, while concerns regarding GenAI exist, they do not universally deter scholarly engagement. Instead, many researchers are actively negotiating these challenges, adopting GenAI with scrutiny, revising their practices, and developing context-sensitive strategies to preserve scholarly standards and research integrity.

#### *4.3.4 What is GenAI to DH researchers?*

This ongoing negotiation is also evident in the diverse attitudes and metaphors researchers use to describe GenAI's role in their work. As illustrated in Table 3, these metaphors range from optimistic engagement to critical skepticism, reflecting a broad spectrum of perspectives.

At one end of the spectrum, "optimistic visionaries" envision transformative applications. As one participant noted, "If you could have a conversation with something trained on 200 years of the *New York Times*, you're probably gonna be able to pull that information in a way keyword searching just doesn't do" (P3). These scholars are eager to see advancements in model tuning and the integration of curated datasets to realize such possibilities. Scholars holding this typical view use metaphors such as "colleague," "research assistant," and "student," to convey a relational and interactive adoption of GenAI as a active collaborative partner in inquiry and interpretation. GenAI in this aspect, is view as an "epistemic technology" as discussed by Alvarado (2023), a system designed for (and inherently capable of) forming knowledge partnerships with human researchers and actively contributing to the intellectual dimensions of their work. Rather than merely executing programmed commands, GenAI has more "agency" to shape research practices, such as participating in brainstorming, offering feedback, and influencing methodological decisions.

Other scholars take a more instrumental or task-oriented perspective in GenAI's role, describing it as a "tool" or a "mathematical predictor." These metaphors reflect a utilitarian stance, in which GenAI is treated as a functional aid that performs specific tasks without deeper engagement in meaning-making. A more ambivalent view emerges in the metaphor of "amplifier," which emphasizes GenAI's capacity to extend both strengths and weaknesses in human thinking. This framing acknowledges the productive potential of GenAI while simultaneously cautioning against its ability to reinforce existing biases or epistemic blind spots.

At the skeptical end of the spectrum, some scholars remain skeptical or resistant to GenAI. They dismiss it as a tool that is "never gonna work… just a fad" (P2) or deliberately avoid incorporating it into their own research (P8). They use metaphors like "dancing bear" to highlight concerns about spectacle and superficiality. Such imagery suggests that GenAI may impress or entertain, but ultimately lacks authentic understanding or critical reasoning–raising doubts about its intellectual seriousness and trustworthiness.

This complex metaphorical landscape reflects a field still in the process of negotiating GenAI's place in scholarly work. The absence of a dominant metaphor or unifying narrative suggests that DH researchers are still experimenting, cautiously exploring, and evaluating the implications of GenAI in ways that resist premature consensus.

Table 3. Metaphors used by interview participants to describe GenAI.

| Metaphor | Quote(s) | Participant(s) |
|---|---|---|
| **Expert Colleague** | "So he treats GPT like a colleague... he's very happy that he has an expert colleague who can help him with these Latin translations." | P2 |
| **Actor** | "Maybe another analogy is you can get an actor to play 5 different roles." | P2 |
| **Amplifier** | "...I would also put it in as it is an amplifier. Amplifier of one enlarges our power... as well as enlarging our bias..." | P22 |
| **Research Assistant** | "Sort of like a research assistant if you hire an undergrad to do some work..." | P8; P19 |
| **Student** | "...GPT is a student…and you have to instruct GPT to help you find the best way…" | P23 |
| **Tool** | "ChatGPT is just another tool in the toolbox." | P3 |
| **Mathematical Predicator** | "It's just a prediction…it's a mathematical representation of a probability and that's it." | P22 |
| **Dancing Bear / Seal** | "They see GPT more as a curiosity, like a dancing bear. Or like a seal that can balance a balloon on its nose." | P2 |

## 5. Discussion: Is GenAI transforming DH research, and how?

### 5.1 DH actor networks and negotiated process of enrolment and translation

Our findings reveal a dynamically evolving actor-network in DH, composed of both human and non-human participants: individual researchers, the broader DH community (colleagues and students), GenAI, research tasks, external support (funding structures, technical support, organizational policies), and disciplinary norms. Each of these actors is enrolled at different points in the research process and assumes a distinct role. For instance, a researcher—shaped by their disciplinary background, methodological training, or any external factor—may elect to deploy GenAI for a particular research task.

Once enrolled, GenAI becomes an actor: researchers may assign it tasks, yet GenAI also informs their decisions, reshapes research practices, and—even reallocates work in some cases. And these interactions form a dynamic actor-network whose configurations range from tightly interdependent to more parallel or loosely connected arrangements. Certain elements—core research tasks or funding structures, for instance—function as obligatory passage points (Latour, 1987), bottlenecks that determine which actor can participate and how, and that help bring actors into alignment around shared questions and purposes. For example, one interviewee (P21) used multimodal GenAI to decipher cursive handwriting in archival work, while another (P27) generated synthetic song lyrics to navigate copyright constraints in cultural analytics. Such decisions are mediated by factors such as peer expectations, the affordances and limitations of GenAI, task specificity, and the availability of external support (technical, financial, or institutional). Figure 6 illustrates the dynamic interactions within this DH actor network, as represented in our survey and interview data.

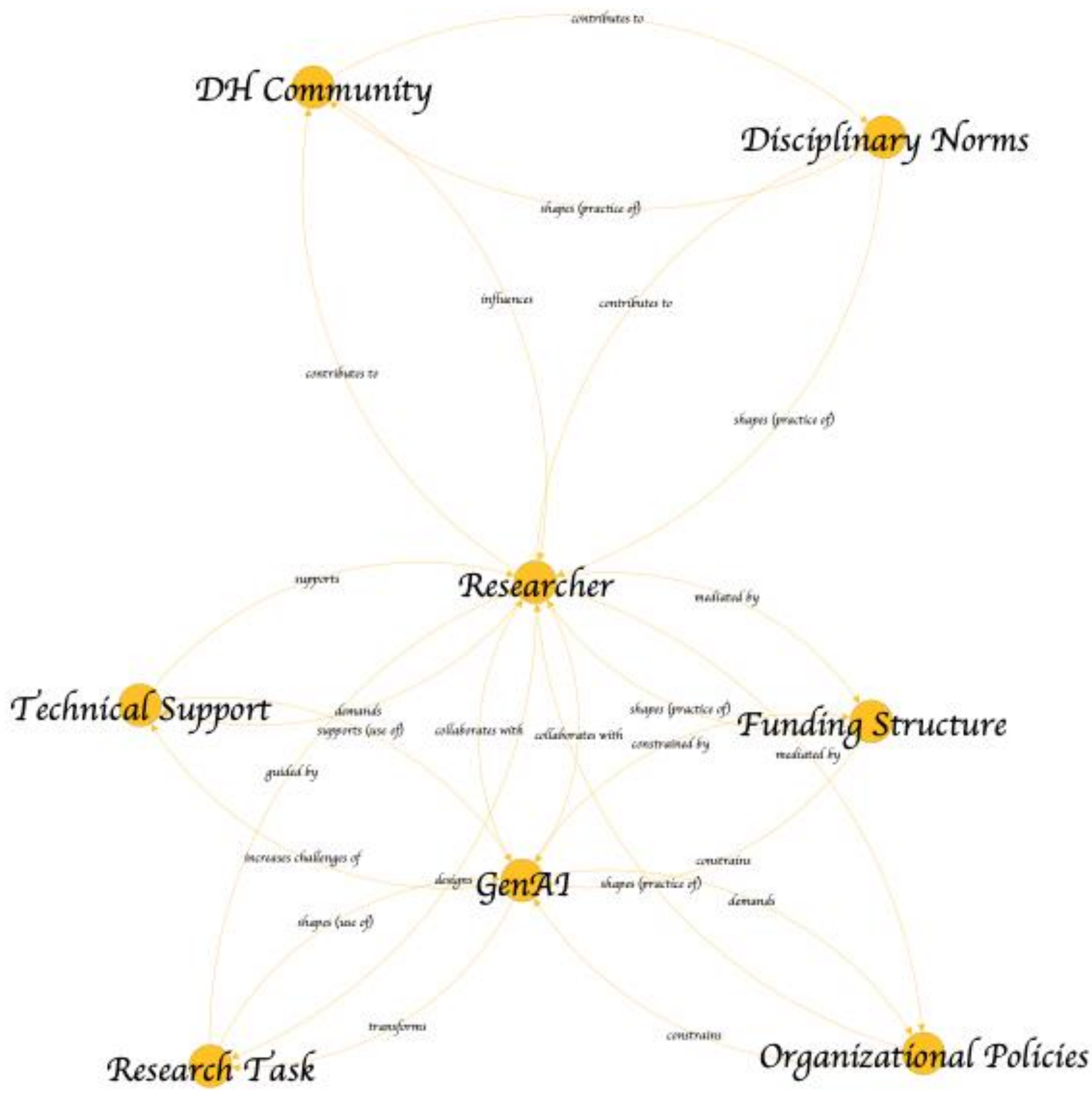

Figure 6. Direct Actor Network of Digital Humanities. Nodes represent actors; edges represent the nature and direction of each relationship between actors.

From an ANT perspective, this emergent, reconfigured network holds the power to transform how DH knowledge is produced. Rather than unfolding through abrupt, top-down paradigm shifts, disciplinary change emerges slowly, through the negotiated realignment of heterogeneous human and non-human actors (Latour, 2005). Scientific and scholarly innovation is thus accumulative—built through successive processes of enrolment, translation, and assembly of all relevant actors and roles.

### 5.2 Nonhuman agency and the roles of GenAI

Our findings also indicate GenAI as a nonhuman actor endowed with "agency." ANT highlights nonhuman agency, and our findings demonstrate that GenAI actively reshapes the constellation of DH actors through negotiated, uneven processes (Sayes, 2014). In Latour's terms, GenAI functions as a mediator that transforms researchers' intentions rather than merely transmitting them. For example, some DH scholars integrate GenAI into core research activities such as multilingual translations, brainstorming and the generation of creative ideas, as well as manuscript editing–tasks that are labor-intensive and specialized. In doing so, GenAI becomes part of the inscription system that converts messy, situated information into integrated, portable forms—what Latour (1990) calls immutable mobiles. As such, GenAI extends beyond being a technical asset and becomes more of a constitutive element of DH intellectual workflows. Its integration into knowledge production blurs the boundary between tool and collaborator, raising new questions about intellectual agency and the distribution of expertise.

These may be attributed to the unique nature of GenAI technologies that differs from earlier digital tools used in DH. Previous waves of DH innovation, such as TEI encoding, GIS, database construction, and even topic modeling, were largely modular and procedural: they extended the capabilities of human researchers but remained firmly under their interpretive control. By contrast, GenAI introduces probabilistic and generative systems that can autonomously synthesize, transform, or produce new content, often in ways that are opaque even to their users. As a result, GenAI tools could reconfigure the cognitive and interpretive terrain of DH, rather than merely augmenting existing methods.

We also observed how DH scholars are actively negotiating the role and identity of GenAI in their work. The variety of metaphors used by participants, such as "research assistant," "colleague," or "student," suggest that GenAI is being enrolled as a semi-autonomous actor, capable of contributing intellectually to tasks such as coding, visualization, and analysis. For instance, P3 described how GenAI changed the allocation of work during research process, facilitating researchers to shift their focus to "think at a higher level;" while P8 used it to do data visualizations that complement the skillset of the researcher. At the same time, other metaphors, such as P2's reference to GenAI as a "dancing bear" or a "seal balancing a ball," signal skepticism about GenAI's epistemic legitimacy and refusal of GenAI as a durable actor within DH networks. The fact that some scholars enthusiastically integrate GenAI while others actively reject it (e.g., P2, P8) highlights the ongoing contestation over its role. The DH actor-network therefore remains unsettled, with GenAI's position continually translated, negotiated, and tested.

The dynamic, unsettled network opens the door to new kinds of epistemic partnerships between DH scholars and GenAI (Alvarado, 2023; Ferrario et al., 2024). Its generative capacities invite engagement in interpretation, creativity, and hypothesis formation, which is central to humanities research. As treat GenAI as a dialogic collaborator rather than a mere tool, they foster hybrid inquiry that merges human judgment with algorithmic reasoning. These evolving partnerships could broaden DH's epistemic repertoire, blur the boundary between tool and thinker, and ultimately redefine interpretation, expertise, and the nature of humanistic scholarship itself (Hauswald, 2025).

In summary, our findings suggest that GenAI can transform DH gradually through translation and negotiation. Rather than replacing existing practices wholesale, GenAI is enrolled, translated, negotiated, and reimagined, within the emergent DH actor network. This aligns with the historical trajectory of DH itself, which, as Dalbello (2011) notes, has always developed through overlapping layers of technological innovation, infrastructural adaptation, and negotiation. However, whether GenAI will stabilize as a core element of this DH actor network will depend on how the negotiations unfold and evolve over time. ANT allows us to see this transformation not as a binary of adoption or rejection, but as a dynamic and ongoing process of relational reconfiguration, one in which epistemic authority, scholarly labor, and disciplinary identity are all in flux.

## 6. Conclusion and Future Work

This study offers one of the first empirical investigations into how GenAI is being adopted, interpreted, and negotiated within the DH research community. Drawing on an international survey of 76 DH scholars and 15 in-depth interviews, we analyzed the diverse rationales behind GenAI adoption, documented specific research practices involving GenAI tools and applications, and examined how scholars perceive the benefits, risks, and implications of this emerging technology. Our findings reveal a DH landscape in flux, marked by both enthusiasm for increased efficiency and new creative affordances as well as concerns over the erosion of intellectual identity, shifts in scholarly labor, and the integrity of core disciplinary values.

From the perspective of ANT, these findings suggest that GenAI is not producing a disruptive paradigm shift in the Kuhnian sense but is gradually reconfiguring the actor networks that constitute DH. We leveraged ANT to show an emergent network within DH, where GenAI is being enroled as a new kind of actor—sometimes as a tool, sometimes as a collaborator—and is increasingly integrated into research workflows that shapes not only outputs, but also intentions, methods, and epistemic norms. As DH scholars begin to assign metaphors like "research assistant," "student," or "colleague" to GenAI, they signal emerging epistemic partnerships that challenge traditional boundaries between interpretation and automation, between human insight and machine-generated output. Importantly, these transformations are uneven, negotiated, and highly contextual. Scholars vary widely in their degree of engagement with GenAI, and the actor-network remains unsettled, with competing visions about what role, if any, GenAI should play in the future of the field. Some participants articulated resistance through distancing metaphors (e.g., "dancing bear"), reflecting deeper anxieties about GenAI's epistemic legitimacy and the potential displacement of core humanistic practices. Others, by contrast, saw GenAI as a catalyst for rethinking their workflows, developing new skills, or imagining alternative research futures.

In this respect, GenAI differs meaningfully from previous technologies adopted within DH. With generative and interpretive capacities that engage directly with the intellectual core of humanistic inquiry, GenAI participates in meaning-making, hypothesis generation, and language production, thereby raising new questions about authorship, credibility of research, and scholarly agency. Because of this distinct capacity, GenAI may reshape epistemic partnerships between DH researchers and technology, and gradually, transform how DH scholars value interpretive labor and rethink their disciplinary essence. Beyond the DH domain, this study also contributes to information science by engaging a longstanding discourse on how emerging information technologies—particularly GenAI—both shape and are shaped by the values, infrastructures, and practices of knowledge-producing communities.

Despite the insights garnered in the current work, we acknowledge certain limitations of ANT. ANT conceptualizes institutions not as pre-existing structures that constrain or enable action, but as outcomes of stabilized associations among human and non-human actors. Institutions, in this view, are continually enacted and maintained through networked practices. While this perspective illuminates the dynamic and contingent nature of institutional formation, ANT's strong focus on localized interactions can obscure the influence of broader, macro-level forces. Regulatory frameworks, legal systems, and economic or cultural structures often shape actor-networks in ways that are difficult to capture through micro-level analysis alone. For example, the enduring impact of DH funding mechanisms or disciplinary boundaries may exert systemic pressures that resist change, which ANT may not fully account for when emphasizing situated negotiations.

Recognizing these limitations, our study highlights the need to complement ANT with frameworks that account for both localized practices and broader institutional contexts. In our current findings, we identified several key social mechanisms that influence how GenAI is negotiated within DH, such as funding availability, institutional support, and the presence (or absence) of formal guidelines and established policies. Although our current dataset does not yet support a systematic analysis of how these broader factors shape GenAI adoption, we plan to pursue this line of inquiry in future research. Specifically, we aim to examine whether the emerging actor-network around GenAI in DH can be sustained or institutionalized over time. To advance this investigation, we propose incorporating theoretical frameworks that focus more on institutions and broader social environments. For example, the Socio-Technical Interaction Network (STIN) framework offers a valuable and actionable social informatics approach for identifying key "invisible actors," i.e., the institutional and social mechanisms, that affect the uptake, embedding, and routinization of GenAI technologies (Lamb et al., 2000; Lamb & Kling, 2003; Meyer, 2006; Sawyer & Tyworth, 2006). By investigating these specific factors in-depth with the STIN approach, we plan to explain how broader institutional, infrastructural, and policy constructs enable or constrain the integration of GenAI into scholarly practice. Additionally, other theoretical lenses, such as communities of practice (Cox, 2005), or information practices (Savolainen, 2008), may provide further insights into the localized, collective, and practice-based dynamics that shape a sustainable adoption of GenAI as well. These approaches will guide future phases of our research as we move toward understanding the long-term institutionalization of GenAI in the field.

In addition, we recognize other limitations of the study and aim to address them in future work. First, the relatively small sample size may introduce sampling bias and limit the generalizability of our findings to the broader DH community. Given that GenAI adoption in DH is still in its early stages, identifying and reaching a wider pool of potential respondents remains challenging. Second, because most of our data were collected in 2024, scholars' practices and perceptions may have shifted alongside the rapid evolution of GenAI tools. To capture these developments, we plan to conduct longitudinal research to investigate how DH actor networks, and particularly scholars' engagement with GenAI, evolve over time, tracking technological advances, expanding institutional support, and the stabilization of community norms. In tandem, we will examine the evolving metaphors and discourses around GenAI within DH communities to better understand how scholars imagine and negotiate its agency.

In conclusion, these efforts aim to build a more comprehensive understanding of GenAI's evolving role in DH. As our findings suggest, the emergence of GenAI is catalyzing a moment of field transformation in DH history. Rather than signaling a rupture, GenAI's influence is unfolding through ongoing acts of translation and negotiation. Whether it ultimately stabilizes as a core actor in the DH knowledge ecosystem will depend not only on technological capacity, but also on the relational, material, and institutional work of scholars who are reimagining what it means to produce knowledge in the emerging AI-driven information age.

## Data Availability Statement

The survey questionnaire, interview protocol, and codebook of this study are provided in the supplementary materials. De-identified survey responses are openly available at: Ma, R., & Dedema, M. (2025). De-identified survey responses for Generative AI in DH Research [Data set]. Zenodo. https://doi.org/10.5281/zenodo.15881361. Due to privacy considerations, interview transcripts are not publicly shared but are available upon request from the corresponding author.

## Acknowledgement

The authors would like to thank all survey respondents and interview participants for generously sharing their insights—this paper would not have been possible without their contributions. We also extend our sincere thanks to Dr. Howard Rosenbaum for his valuable feedback on the theoretical frameworks during the development of the manuscript.

## References

Andersen, J. P., Degn, L., Fishberg, R., Graversen, E. K., Horbach, S. P. J. M., Schmidt, E. K., Schneider, J. W., & Sørensen, M. P. (2025). Generative Artificial Intelligence (GenAI) in the research process – A survey of researchers' practices and perceptions. *Technology in Society*, *81*, 102813. https://doi.org/10.1016/j.techsoc.2025.102813

Alvarado, R. (2023). AI as an Epistemic Technology. *Science and Engineering Ethics*, *29*(5), 32.

Berry, D. M. (2022). *AI, Ethics, and Digital Humanities*. In O'Sullivan, J (Ed.), The Bloomsbury Handbook to the Digital Humanities, Bloomsbury Academic.

Berthelot, A., Caron, E., Jay, M., & Lefèvre, L. (2025). Understanding the environmental impact of generative AI services. *Commun. ACM*, *68*(7), 46–53. https://doi.org/10.1145/3725984

Buruk, O. "Oz." (2023). Academic writing with GPT-3.5 (ChatGPT): Reflections on practices, efficacy and transparency. *26th International Academic Mindtrek Conference*, 144–153. https://doi.org/10.1145/3616961.3616992

Charmaz, K. (2006). *Constructing grounded theory: a practical guide through qualitative analysis.* London: SAGE.

Colutto, S., Kahle, P., Hackl, G., & Mühlberger, G. (2019). Transkribus: A platform for automated text recognition and searching of historical documents. In *2019 15th International Conference on eScience (eScience)*, 463–466). IEEE. https://doi.org/10.1109/eScience.2019.00060.

Cottom, T. (2016). More scale, more questions: Observations from sociology. In M. K. Gold & L. F. Klein (Eds.), *Debates in the Digital Humanities 2016*. University of Minnesota Press.

Cox, A. (2005). What are communities of practice? A comparative review of four seminal works. *Journal of Information Science, 31*(6), 527–540. https://doi.org/10.1177/0165551505057016

Creswell, J. W., & Creswell, J. D. (2023). *Research design: Qualitative, quantitative, and mixed methods approaches.* SAGE Publications, Inc.

Dalbello, M. (2011). A genealogy of digital humanities. *Journal of documentation, 67*(3), 480–506.

Deegan, M. (2013). 'This ever more amorphous thing called Digital Humanities': Whither the Humanities Project? *Arts and Humanities in Higher Education*, *13*(1-2), 24–41. https://doi.org/10.1177/1474022213513180

Delios, A., Tung, R. L., & van Witteloostuijn, A. (2025). How to intelligently embrace generative AI: The first guardrails for the use of GenAI in IB research. *Journal of International Business Studies*, *56*(4), 451–460. https://doi.org/10.1057/s41267-024-00736-0

Doron, G., Genway, S., Roberts, M., & Jasti, S. (2025). Generative AI: Driving productivity and scientific breakthroughs in pharmaceutical R&D. *Drug Discovery Today*, *30*(1), 104272. https://doi.org/10.1016/j.drudis.2024.104272

Dwivedi, Y. K., Kshetri, N., Hughes, L., Slade, E. L., Jeyaraj, A., Kar, A. K., Baabdullah, A. M., Koohang, A., Raghavan, V., Ahuja, M., Albanna, H., Albashrawi, M. A., Al-Busaidi, A. S., Balakrishnan, J., Barlette, Y., Basu, S., Bose, I., Brooks, L., Buhalis, D., … Wright, R. (2023). Opinion Paper: "So what if ChatGPT wrote it?" Multidisciplinary perspectives on opportunities, challenges and implications of generative conversational AI for research, practice and policy. *International Journal of Information Management*, *71*, 102642. https://doi.org/10.1016/j.ijinfomgt.2023.102642

Ferrario, A., Facchini, A., & Termine, A. (2024). Experts or authorities? The strange case of the presumed epistemic superiority of artificial intelligence systems. *Minds and Machines*, *34*(3), 30.

Filimonovic, D., Rutzer, C., & Wunsch, C. (2025). Can GenAI improve academic performance? Evidence from the social and behavioral sciences. *Social Science Research Network* (SSRN Scholarly Paper 5225360). https://doi.org/10.2139/ssrn.5225360

Gefen, A., Saint-Raymond, L., & Venturini, T. (2021). AI for digital humanities and computational social sciences. *Reflections on Artificial Intelligence for Humanity*, 191–202. https://doi.org/10.1007/978-3-030-69128-8_12

Given, L. M., & Willson, R. (2018). Information technology and the humanities scholar: Documenting digital research practices. *Journal of the Association for Information Science and Technology, 69*(6), 807–819.

Guo, Q. (2024). Prompting change: ChatGPT's impact on digital humanities pedagogy – A case study in art history. *International Journal of Humanities and Arts Computing*, *18*(1), 58–78. https://doi.org/10.3366/ijhac.2024.0321

Hauswald, R. (2025). Artificial epistemic authorities. *Social Epistemology*, 1–10.

Jockers, M. (2013). *Macroanalysis: digital methods and literary history.* University of Illinois Press.

Karjus, A. (2023). Machine-assisted mixed methods: Augmenting humanities and social sciences with artificial intelligence. *arXiv preprint*. https://doi.org/10.48550/arXiv.2309.14379

Kennan, M. A., & Cecez-Kecmanovic, D. (2007). Reassembling scholarly publishing: Institutional repositories, open access, and the process of change. In *Proceedings of the 18th Australasian Conference on Information Systems (ACIS)*, 799–808.

Khlaif, Z. N., Mousa, A., Hattab, M. K., Itmazi, J., Hassan, A. A., Sanmugam, M., & Ayyoub, A. (2023). The potential and concerns of using AI in scientific research: ChatGPT performance evaluation. *JMIR Medical Education*, *9*, e47049. https://doi.org/10.2196/47049

Kuhn, T. S. (1962). *The structure of scientific revolutions.* University of Chicago Press.

Kwon, D. (2025). Is it OK for AI to write science papers? Nature survey shows researchers are split. *Nature*, *641*(8063), 574-578. https://www.nature.com/articles/d41586-025-01463-8

Lamb, R., & Kling, R. (2003). Reconceptualizing Users as Social Actors in Information Systems Research. *MIS Quarterly*, *27*(2), 197–236. https://doi.org/10.2307/30036529

Lamb, R., Sawyer, S., & Kling, R. (2000). A Social Informatics Perspective on Socio-Technical Networks. In H. M. Chung (Ed.), *Proceedings ofthe Americas Conference on Information Systems*. Long Beach, CA.

Latour, B. (1987). *Science in action: How to follow scientists and engineers through society.* Harvard University Press.

Latour, B. (1988). A relativist account of Einstein's relativity. *Social Studies of Science, 18*, 3–44.

Latour, B. (1990). Drawing things together. In M. Lynch & S. Woolgar (Eds.), *Representation in scientific practice*. The MIT Press.

Latour, B. (1992). Where are the missing masses? The sociology of a few mundane artifacts. In: W. E. Bijker & J. Law (Eds.), *Shaping Technology/Building Society: Studies in sociotechnical change* (pp. 225–258). Cambridge, MA: MIT Press.

Latour, B. (1996). On actor-network theory: A few clarifications. *Soziale Welt*, *47*(4), 369–381. Retrieved from https://www.jstor.org/stable/40878163.

Latour, B. (2005). *Reassembling the social: An introduction to Actor-Network-Theory*. Oxford University Press.

Luhmann, J., & Burghardt, M. (2022). Digital humanities—A discipline in its own right? An analysis of the role and position of digital humanities in the academic landscape. *Journal of the Association for Information Science and Technology*, *72*(12), 1475–1498. https://doi.org/10.1002/asi.24533

Lund, B. D., Wang, T., Mannuru, N. R., Nie, B., Shimray, S., & Wang, Z. (2023). ChatGPT and a new academic reality: Artificial intelligence-written research papers and the ethics of the large language models in scholarly publishing. *Journal of the Association for Information Science and Technology*, *74*(5), 570–581. https://doi.org/10.1002/asi.24750

Luo, W. (2020). *Translation as actor-networking: Actors, agencies, and networks in the making of Arthur Waley's English translation of the Chinese Journey to the West.* Routledge.

Marchese, D. (2022). An A.I. pioneer on what we should really fear. *The New York Times.* Retrieved from: https://www.nytimes.com/interactive/2022/12/26/magazine/yejin-choi-interview.html

Merton, R. (1973). *The sociology of science: Theoretical and empirical investigations.* The University of Chicago Press.

Meyer, E.T. (2006). Socio-Technical Interaction Networks: A Discussion of the Strengths, Weaknesses and Future of Kling's STIN Model. In: Berleur, J., Nurminen, M.I., Impagliazzo, J. (eds) *Social Informatics: An Information Society for all? In Remembrance of Rob Kling. HCC 2006. IFIP International Federation for Information Processing, vol 223*. Springer, Boston, MA.

Mitchell, M. (2023). AI's challenge of understanding the world. *Science* 382, eadm8175. https://doi.org/10.1126/science.adm8175

Moretti, F. (2013). *Distant reading.* Verso.

Morland, C., & Pettersen, I. (2018). Translating technological change–Implementing technology into a hospital. *International Journal of Productivity and Performance Management, 67*(9), 1786–1800. https://doi.org/10.1108/IJPPM-10-2017-0250

Murdoch, J. (1997). Inhuman/nonhuman/human: Actor-network theory and the prospects for a nondualistic and symmetrical perspective on nature and society. *Environment and Planning D: Society and Space, 15*(6), 731–756.

Murugesan, S. (2025). The rise of agentic AI: Implications, concerns, and the path forward. *IEEE Intelligent Systems, 40*(2), 8–14. https://doi.org/10.1109/MIS.2025.3544940

Nockels, J., Gooding, P., Ames, S., & Terras, M. (2022). Understanding the application of handwritten text recognition technology in heritage contexts: A systematic review of Transkribus in published research. *Archival Science, 22*(3), 367–392. https://doi.org/10.1007/s10502-022-09397-0

Oxtoby, D. W., Allen, D., & Brown, J. (2023). The humanities and the rise of the terabytes. *Bulletin of the American Academy of Arts and Sciences*, *76*(3), 30–41. https://www.jstor.org/stable/10.2307/27216739

Poole, A. H. (2017). The conceptual ecology of digital humanities. *Journal of Documentation, 73*(1), 91–122.

Rane, N. (2023). Role and challenges of ChatGPT and similar generative artificial intelligence in arts and humanities. Available at SSRN: http://dx.doi.org/10.2139/ssrn.4603208

Reif, J. A., Larrick, R. P., & Soll, J. B. (2025). Evidence of a social evaluation penalty for using AI. *Proceedings of the National Academy of Sciences, 122*(19), e2426766122. https://doi.org/10.1073/pnas.2426766122

Savolainen, R. (2008). *Everyday information practices: A social phenomenological perspective*. Scarecrow Press.

Sawyer, S., Tyworth, M. (2006). Social Informatics: Principles, Theory, and Practice. In: Berleur, J., Nurminen, M.I., Impagliazzo, J. (eds) *Social Informatics: An Information Society for all? In Remembrance of Rob Kling. HCC 2006. IFIP International Federation for Information Processing, vol 223*. Springer, Boston, MA.

Sayes, E. (2014). Actor-network theory and methodology: Just what does it mean to say that nonhumans have agency? *Social Studies of Science, 44*(1), 134–149. https://www.jstor.org/stable/43284223

Schlagwein, D., & Willcocks, L. (2023). 'ChatGPT et al.': The ethics of using (generative) artificial intelligence in research and science. *Journal of Information Technology*, *38*(3), 232–238. https://doi.org/10.1177/02683962231200411

Schmidt, B. (2016). Do digital humanists need to understand algorithms? In M. K. Gold & L. F. Klein (Eds.), *Debates in the Digital Humanities 2016*. University of Minnesota Press.

Sinha, R., Solola, I., Nguyen, H., Swanson, H., & Lawrence, L. (2024). The role of generative AI in qualitative research: GPT-4's contributions to a grounded theory analysis. In *Proceedings of the Symposium on Learning, Design and Technology*, 17–25. https://doi.org/10.1145/3663433.3663456

Smithies, J. (2017). *The digital humanities and the digital modern.* Palgrave Macmillan.

Svensson, P. (2009). Humanities computing as digital humanities. *Digital Humanities Quarterly, 3*(3). http://digitalhumanities.org/dhq/vol/3/3/000065/000065.html

Svensson, P. (2010). The landscape of digital humanities. *Digital Humanities Quarterly, 4*(1). http://digitalhumanities.org/dhq/vol/4/1/000080/000080.html

Underwood, T. (2016). Distant reading and recent intellectual history. In M. K. Gold & L. F. Klein (Eds.), *Debates in the Digital Humanities 2016*. University of Minnesota Press.

Underwood, T. (2019). *Distant horizons: Digital evidence and literary change*. University of Chicago Press. https://doi.org/10.7208/9780226612973

Underwood, T. (2025). The impact of language models on the humanities and vice versa. *Nature Computational Science*. https://doi.org/10.1038/s43588-025-00819-4

UNESCO. (2021). Recommendation on the ethics of Artificial Intelligence. (Programme Document SHS/BIO/PI/2021/1). Retrieved from the United Nations Educational, Scientific and Cultural Organization website. https://unesdoc.unesco.org/ark:/48223/pf0000381137

Van Noorden, R., & Perkel, J. M. (2023). AI and science: What 1,600 researchers think. *Nature*, *621*(7980), 672–675. https://doi.org/10.1038/d41586-023-02980-0

Walsh, J. A., Cobb, P. J., de Fremery, W., Golub, K., Keah, H., Kim, J., Kiplang'at, J., Liu, Y.-H., Mahony, S., Oh, S. G., Sula, C. A., Underwood, T., & Wang, X. (2021). Digital humanities in the iSchool. *Journal of the Association for Information Science and Technology*, 73(2), 188–203. https://doi.org/10.1002/asi.24535

Wang, C., Boerman, S. C., Kroon, A. C., Möller, J., & H de Vreese, C. (2024). The artificial intelligence divide: Who is the most vulnerable? *New Media & Society*, 14614448241232345. https://doi.org/10.1177/14614448241232345

Weingart, S. B., & Eichmann-Kalwara, N. (2017). What's under the big tent?: A study of ADHO conference abstracts. *Digital Studies*, *7*(1), 6. https://doi.org/10.16995/dscn.284

Zeng, M.L., Sula, C.A., Gracy, K.F., Hyvönen, E. and Alves Lima, V.M. (2022), *JASIST* special issue on digital humanities (DH). J Assoc Inf Sci Technol, 73: 143-147. https://doi.org/10.1002/asi.24584

Zhu, J., & Molnar, A. (2025). The end of writing as we know it? Generative AI may undermine the social signaling function of writing [Preprint]. https://osf.io/preprints/psyarxiv/j8zgv_v1